# Symbolic Priors for RNN-based Semantic Parsing

**Chunyang Xiao**  **Marc Dymetman**
Xerox Research Centre Europe
chunyang.xiao,marc.dymetman@xerox.com

**Claire Gardent**
CNRS, LORIA, UMR 7503
claire.gardent@loria.fr

## Abstract

Seq2seq models based on Recurrent Neural Networks (RNNs) have recently received a lot of attention in the domain of Semantic Parsing for Question Answering. While in principle they can be trained directly on pairs (natural language utterances, logical forms), their performance is limited by the amount of available data. To alleviate this problem, we propose to exploit various sources of prior knowledge: the well-formedness of the logical forms is modeled by a weighted context-free grammar; the likelihood that certain entities present in the input utterance are also present in the logical form is modeled by weighted finite-state automata. The grammar and automata are combined together through an efficient intersection algorithm to form a soft guide ("background") to the RNN. We test our method on an extension of the *Overnight* dataset and show that it not only strongly improves over an RNN baseline, but also outperforms non-RNN models based on rich sets of hand-crafted features.

## 1 Introduction

Building a Question Answering system has received a lot of attention in recent years [Berant *et al.*, 2013; Reddy *et al.*, 2014; Pasupat and Liang, 2015; Neelakantan *et al.*, 2016]. The key component in such systems is a semantic parser which maps natural language utterances (NLs) to executable logical forms (LFs). Traditional ways of building such semantic parsers [Kwiatkowski *et al.*, 2013; Berant and Liang, 2014; Reddy *et al.*, 2014; Pasupat and Liang, 2015] exploit rich prior knowledge in the form of features and grammars. This type of knowledge facilitates generalization on test data but often fails to adapt to the actual regularities present in the data. For example, Reddy et al. [2014] propose to build a semantic parser based on Combinatorial Categorical Grammar (CCG) [Clark and Curran, 2007] and achieve good results; however, for the WebQuestions dataset [Berant *et al.*, 2013], the authors observe that 25% of system failures are due to the fact that the sentence cannot be parsed by the CCG.

While it is difficult to manually design appropriate features for a specific dataset, one would like a system to discover those features automatically given enough data. Recent semantic parsing systems explore this direction and propose to use recurrent neural networks (RNNs) and particularly LSTMs [Hochreiter and Schmidhuber, 1997] based deep learning models [Jia and Liang, 2016; Dong and Lapata, 2016; Xiao *et al.*, 2016; Neelakantan *et al.*, 2016]; although being already quite successful, those deep learning models can be further enhanced by incorporating prior knowledge and this paper builds on two recent attempts to *combine* prior knowledge with neural networks in an NLP context.

In the context of Natural Language Generation (NLG), Goyal et al. [2016] describe an RNN model that generates sentences character-by-character, conditional on a semantic input. They use a form of prior knowledge, which they call a "background", to guide the RNN in producing string of characters which are (i) valid common English words or (ii) "named entities" (e.g. hotel names, addresses, phone numbers, ...) for which evidence can be found in the semantic input.

In the context of Semantic Parsing, Xiao et al. [2016] propose to use an RNN-based model to predict derivation sequences (DS) that are derivation steps relative to an *a priori* given underlying grammar. The grammar is used to incrementally filter out those derivation steps that may lead to non-interpretable LFs, something which is difficult for the RNN to learn on it own.

While the "background" used by [Goyal *et al.*, 2016] is partially based on its actual semantic input, the prior employed by [Xiao *et al.*, 2016] only exploits knowledge about output well-formedness. In both cases (NLG and Semantic Parsing) however, the output depends on the input; In semantic parsing, if the input question contains the string 'Barack Obama', it is highly likely that the LF of that question involves the entity Barack Obama and therefore, that the rule expanding to "barack obama" is present in the output derivation sequence.

This work can be seen as an extension of the semantic parsing approach proposed in [Xiao *et al.*, 2016] using ideas from [Goyal *et al.*, 2016], where we use a background prior that combines the grammaticality constraints of [Xiao *et al.*, 2016] with certain types of prior beliefs that we can extract from the NL question.

Combining different sources of prior knowledge, which can also be seen as combining different factors in a graphical model, is a hard problem. In general, to compute the exact combination (with even two factors), one does not have other solutions than to go through an exhaustive enumeration of both factors and multiplying each pair of factors. Our proposed solution to this problem is to implement our input-dependent background through weighted finite-state automata (WFSAs), which we then *intersect* with a WCFG representing valid grammar derivations.

Intersecting a WFSA with a WCFG can be done through a dynamic programming procedure (thus efficient as it avoids exhaustive enumeration) closely related to chart-parsing. The result of this intersection algorithm is a new WCFG, which can be normalized into a PCFG (Probabilistic CFG), which makes explicit the conditional probabilities for the different ways in which a given derivation sequence can be continued.[1]

The obtained PCFG is then used as a background, and when making its next local choice, the RNN has only to learn to "correct" the choices of the PCFG. In the cases where the background is close to the true distribution, the RNN will learn to predict a uniform distribution thus always referring to the background for such predictions.

This is in fact a desirable behaviour as the background may contain prior knowledge that the RNN is not able to learn based on data (e.g. prior knowledge on entities unseen in training) and the best behavior for the model in those cases is to refer to the background.

We test our new Background RNN semantic parser on an extended version of the *Overnight* dataset [Wang *et al.*, 2015], which removes certain problematic aspects of the original dataset (that made the results too optimistic, as explained in section 4). By incorporating simple input-dependent prior knowledge via WFSAs, our model not only improves over its RNN baseline but also over the non-RNN system proposed in [Wang *et al.*, 2015] which involves much richer hand-crafted features.

## 2 Paper Background

### 2.1 The Approach of Wang et al. (2015), Original and New Datasets

```
s0:        s(S) → np(S).
np0:       np(get[CP,NP]) → np(NP), cp(CP).
np1:       np(NP) → typenp(NP).
cp0:       cp([lambda,s,[filter,s,RELNP,=,ENTNP]]) →
               [whose], relnp(RELNP), [is], entitynp(ENTNP).
...
typenp0:   typenp(article) → [article].
relnp0:    relnp(pubDate) → [publication, date]
entitynp0: entitynp(1950) → [1950].
...
```

Figure 1: Some general rules (top) and domain-specific rules (bottom) of the *Overnight* in DCG format.

[Wang *et al.*, 2015] (which we refer to as "SPO") proposes a novel way to build datasets for training a semantic parser without having to manually annotate natural language sentences with logical forms (LFs). First a grammar (of which we provide an extract in Fig. 1, in the format of Definite Clause Grammars [Pereira and Warren, 1980], reproduced from [Xiao *et al.*, 2016]) is used to generate LFs paired with conventional surface realizations called "canonical forms" (CFs). For example, the rules shown in Fig. 1 support the generation of the LF *get[[lambda,s,[filter,s,pubDate,=,1950]],article]* along with the CF "article whose publication date is 1950".

The CFs are not necessarily natural English but are supposed to be "semantically transparent" so that one can use crowdsourcing to paraphrase those CFs into natural utterances (NLs) e.g., *Articles published in 1950*. The resulting (NL, LF) pairs make up a dataset which can be used for learning semantic parsers.

After collecting all the paraphrases, the authors of SPO construct a dataset divided into training and test sets by performing a random 80%-20% split over all the (NL, LF) pairs. However, given the data collecting process, each LF tends to correspond to several (in general more than 5) paraphrases. In consequence, inside this original dataset, most of the LFs in the test set have already been seen in training, making the task close to a classification process and easier than it should be.

In addition, as pointed out by Jia and Liang [2016], the original dataset contains very few named entities. In this work, we therefore construct a new dataset called *Overnight+* fixing some of the above issues. More details on our proposed dataset can be found in section 4.1.

### 2.2 The Approach of Xiao et al. (2016)

To learn the semantic parser, SPO first trains a log-linear model based on rich prior features dependent jointly on NL and the corresponding (LF, CF) pair. Then it searches for the derivation tree relative to the grammar for which the produced (LF, CF) pair has the highest score [Pasupat and Liang, 2015].

In contrast, Xiao et al. [2016] propose to use RNN-based models to directly map the NL to its corresponding derivation sequence (DS). Derivation sequences are sequentialized representations of derivation trees in the grammar. For example, the derivation tree generating the CF "article whose publication date is 1950" is *s0(np0(np1(typenp0),cp0(relnp0,entitynp0)))*; The associated DS is the leftmost traversal of this tree: *s0,np0,np1,typenp0,cp0,relnp0,entitynp0*.

Predicting DS provides an efficient sequentialization and makes it easy to guarantee the well-formedness of the predicted sequence. Xiao et al. [2016] show that their model "Derivation Sequence Predictor with Constraints Loss" (DSP-CL) achieves good performance on the original *Overnight* dataset.

Our work here can be seen as extending DSP-CL by integrating some input-dependent prior knowledge into the RNN predictor, allowing it to improve its performance on the more challenging *Overnight+* dataset.

---

[1] While on first sight the whole procedure may appear somewhat involved, it has the crucial advantage that a global constraint, for instance the required appearance of a certain symbol at some unknown *future* point in the DS, has local consequences much earlier in the incremental process that the network is following.

## 3 Model

### 3.1 Background Priors on RNNs

[Goyal *et al.*, 2016], in the context of NLG, proposes to modify the standard generative procedure of RNNs:

$$p_\theta(x_{t+1}|x_1,\ldots,x_t,C) = rnn_\theta(x_{t+1}|x_1,\ldots,x_t,C),$$

where $C$ is the observed input context (that is, the input of the seq2seq model), $x_1,\ldots,x_t$ the current output prefix, $rnn_\theta$ the softmax output of the RNN parametrized by $\theta$, and $p_\theta$ the probability distribution from which the next symbol $x_{t+1}$ is sampled, with:

$$p_\theta(x_{t+1}|x_1,\ldots,x_t,C) \propto b(x_{t+1}|x_1,\ldots,x_t,C) \\ \cdot rnn_\theta(x_{t+1}|x_1,\ldots,x_t,C), \quad (1)$$

where the *background* $b$ is an arbitrary non-negative function over $C, x_1, \ldots, x_t, x_{t+1}$, which is used to incorporate prior knowledge about the generative process $p_\theta$.[2] On one extreme, taking $b$ to be uniform corresponds to the situation where no prior knowledge is available, and one is back to a standard RNN, with all the discriminating effort falling on the *rnn* component and relying on whatever (possibly limited) training data is available in each context; on the other extreme, if the true process $p$ is known, one may take $b = p$, and then the *rnn* component $rnn_\theta(x_{t+1}|x_1,\ldots,x_t,C)$ is only required to produce a close-to-uniform distribution over the target vocabulary, independently of $x_1,\ldots,x_t,C$, which only requires the layer just before the softmax to produce a close-to-null vector, an easy task to learn (by assigning close-to-null values to some matrices and biases). In practice, the interesting cases fall between these two extremes, with the background $b$ incorporating some prior knowledge that the *rnn* component can leverage in order to more easily fit the training data. In the NLG application considered by [Goyal *et al.*, 2016], the output of the seq2seq model is a string of characters, and the background — implemented as a WFSA over characters — is used to guide this output: (i) towards the production of valid common English words, and (ii) towards the production of named entities (e.g. hotel names, addresses, ...) for which evidence can be found in the semantic input.

The approach of [Xiao *et al.*, 2016] can be reformulated into such a "Background-RNN" (BRNN) framework. In that work, the underlying grammar $G$ acts as a yes-no filter on the incremental proposals of the RNN, and this filtering process guarantees that the evolving DS prefix always remains valid relative to $G$. There:

$$p_\theta(x_{t+1}|x_1,\ldots,x_t,C) \propto b(x_{t+1}|x_1,\ldots,x_t) \\ \cdot rnn_\theta(x_{t+1}|x_1,\ldots,x_t,C), \quad (2)$$

where $C = NL$ is the input question, the $x_i$'s are rule-names, and $b$ takes a value in $\{0, 1\}$, with $b(x_{t+1}|x_1,\ldots,x_t) = 1$ indicating that $x_1,\ldots,x_t,x_{t+1}$ is a valid DS prefix relative to $G$. With this mechanism in place, on the one hand the BRNN cannot produce "ungrammatical" (in the sense of being valid

---

[2]See also [Dymetman and Xiao, 2016], for a more general presentation, of which the background-RNN can be seen as a special case.

according to the grammar) prefixes, and on the other hand it can exploit this grammatical knowledge in order to ease the learning task for the *rnn* component, which is not responsible for detecting ungrammaticality on its own anymore.

### 3.2 WCFG Background

While the (implicit) background of [Xiao *et al.*, 2016] shown in (2) is a binary function that does not depend on the NL input, but only on hard grammaticality judgments, in this paper, we propose to use the more general formulation (1). Now $b$ is soft rather than hard, and it does exploit the NL input.

More specifically, $b(x_{t+1}|x_1,\ldots,x_t,NL)$ is obtained in the following way. First, we use the original grammar $G$ together with the input $NL$ to determine a WCFG (weighted context-free grammar) $GW_{NL}$ over derivation sequences of the original CFG (that is, the terminals of $GW_{NL}$ are rule-names of the original $G$), as will be explained below. Second, we compute $b(x_{t+1}|x_1,\ldots,x_t,NL)$ as the conditional probability relative to $GW_{NL}$ of producing $x_{t+1}$ in the context of the prefix $x_1,\ldots,x_t$.

Apart from our use of a much richer background than [Xiao *et al.*, 2016], our overall training approach remains similar to theirs. Our training set consists in pairs of the form $(NL, DS)$; the $rnn_\theta$ component of the BRNN (1) is a seq2seq LSTM-based network in which the input encoding is a vector based on the unigrams and bigrams present in $NL$, and where the $DS$ output is a sequence of rule-names from $G$; the logit output layer of this network is then combined additively with $\log b$ before a softmax is applied, resulting in the probability distribution $p_\theta(x_{t+1}|x_1,\ldots,x_t,NL)$; finally the incremental cross-entropy loss of the network $-\log p_\theta(\bar{x}_{t+1}|x_1,\ldots,x_t,NL)$ is backpropagated through the network (where $\bar{x}_{t+1}$ is the observation in the training data). Implementation details are provided in section 4.

**Constructing the WCFG background, WFSA factors**

As in the case of [Xiao *et al.*, 2016], we still want our background to ensure grammaticality of the evolving derivation sequences, but in addition we wish it to reflect certain tendencies of these sequences that may depend on the NL input. By stating these tendencies through a real-weighted, rather than binary, background $b(x_{t+1}|\ldots)$, we make it possible for the *rnn* component to bypass the background preferences in the presence of a training observation $x_{t+1}$ that does not agree with them, through giving a high enough value to $rnn_\theta(x_{t+1}|\ldots)$.

Our approach is then the following. We start by constructing a simple WCFG $GW_0$ that (1) enumerates exactly the set of all valid derivation sequences relative to the original $G$, and (2) gives equal weight $1/n_{NT}$ to each of the possible $n_{NT}$ expansions of each of its non-terminals $NT$. Thus $GW_0$ is actually a *Probabilistic* CFG (PCFG), that is, a WCFG that has the property that the sum of weights of the possible rules expanding a given nonterminal is 1. Thus $GW_0$ basically ensures that the strings of symbols it produces are valid DS's relative to $G$, but is otherwise non-committal concerning different ways of extending each prefix.

The second step consists in constructing, possibly based on the input $NL$, a small number of WFSA's (weighted FSA's)

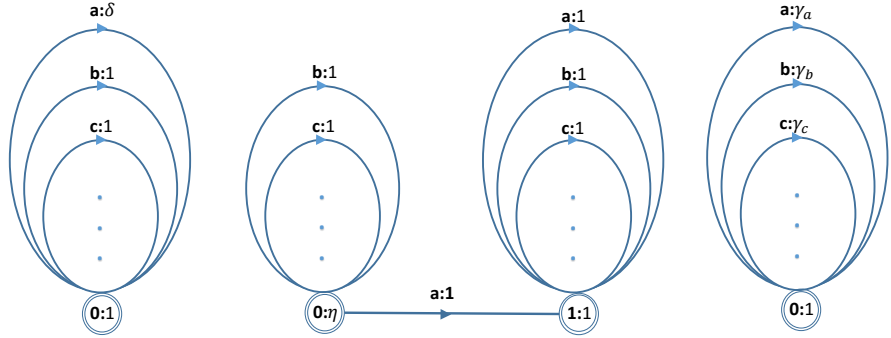

Figure 2: Three WFSA's for handling different types of prior information. Edge labels are written in the form *symbol : weight*. The initial state is **0**. Final states are indicated by a double circle and their exit weight is also indicated.

each of which represents a certain aspect of the prior knowledge we have about the likely output sequences. These automata will be considered as "factors" (in the sense of probabilistic graphical models) that will be intersected (in other words, multiplied) with $GW_0$, resulting in a final WCFG $GW_{NL}$ which will then combine the different aspects and be used as our background.[3] In Fig. 2, we illustrate three possible such automata; here the output vocabulary consists of the symbols $a, b, c, \ldots$ (in our specific case, they will actually be DS symbols, but here we keep the description generic).

Let us first describe the automaton on the left. Here each symbol appears on an edge with weight 1, with the exception of the edge associated with $a$, which carries a weight $\delta \ll 1$; this automaton thus gives a "neutral" weight 1 to any symbol sequence that does not contain $a$, but a much smaller weight $\delta^k$ to one that contains $k \geq 1$ instances of $a$. Once intersected with $GW_0$, this automaton can be used to express the belief that given a certain input $NL$, $a$ is unlikely to appear in the output.

The automaton in the middle expresses the opposite belief. Here the exit weight associated with the final (also initial) state **0** is $\eta \ll 1$. This automaton gives a weight 1 to any sequence that contains $a$, but a weight $\eta$ to sequences that do not. Once intersected with $GW_0$, this automaton expresses the belief that given the input $NL$, $a$ is likely to appear in the output.

The automaton on the right is a simple illustration of the kind of prior beliefs that could be expressed on output sequences, independently of the input. Here $\gamma_x$ denotes the unigram probability of the output symbol $x$. In the context of semantic parsing, such automata on the output could be used to express certain forms of regularities on expected logical forms, such as, like here, unigram probabilities that are not handled by the grammar $GW_0$ (which is concerned only by well-formedness constraints), or more generally, observations about certain patterns that are likely or unlikely to occur in the logical forms (e.g. the unlikeliness of mixing basketball players with scientific authors), insofar as such regularities can be reasonably expressed in finite-state terms.

**Why automata?** In order to be effective, the background $b$ has to be able to provide the *rnn* component with useful information on the *next incremental step*, conditional on the already generated prefix. In addition, we would like the background to capitalize on different sources of information.

In connection with these desiderata, WCFG and WFSAs have the following remarkable properties: (1) the intersection of several WFSAs is a WFSA which can be efficiently computed, (2) the intersection of a WCFG with a WFSA is a WCFG which can be efficiently computed, (3) given a prefix, the conditional probability of the next symbol relative to a WCFG (resp. a WFSA) can be efficiently computed; here "efficiently computed" means through Dynamic Programming and in polynomial time [Nederhof and Satta, 2003]. These properties are conspicuously absent from most other generative devices. For instance it is far from obvious how to intersect two different RNNs to compute the conditional probability of the next symbol, given a common prefix: while a certain symbol may have a large probability relative to *both* RNNs, the later (global) consequences of choosing this next symbol may be largely incompatible between the two RNNs; in other words, the *local* combined conditional probability cannot be computed solely on the basis of the product of the two *local* conditional probabilities.

*Implementation principles.* The fact that one can intersect a WCFG with a WFSA to produce another WCFG is a generalization of the classical result [Bar-Hillel *et al.*, 1961] concerning the non-weighted case. The implementation we use is based on the Earley-inspired intersection algorithm of [Dyer, 2010], obtaining a certain WCFG, which we normalize into probabilistic form [Nederhof and Satta, 2003], finally obtaining a PCFG $GW_{NL}$. In order to compute the background $b(x_{t+1}|x_1, \ldots, x_t, NL)$ we then need to compute the conditional probability relative to $GW_{NL}$ of producing the symbol $x_{t+1}$ given the prefix $x_1, \ldots, x_t$. There are some special-purpose algorithms for doing that efficiently, for instance [Stolcke, 1994], but in this work we use again (unoptimally) the generic Earley intersection algorithm, taking advantage of the fact that the probability mass relative to $GW_{NL}$ of the set of sequences starting with the prefix $x_1, \ldots, x_t, x_{t+1}$ can be obtained by intersecting $GW_{NL}$ with the automaton generating the language of all sequences starting with this prefix.

---

[3]Formally, intersecting a WCFG $GW$ with a WFSA $A$ consists in applying a Dynamic Programming algorithm that constructs a new WCFG $GW' = GW \cap A$. If the weight of a certain sequence $x_1, \ldots, x_n$ is $\omega_{GW}$ relative to $GW$ (resp. $\omega_A$ relative to $A$), then its weight relative to $GW'$ is $\omega_{GW} \cdot \omega_A$.

## 4 Experiments

### 4.1 Setup

The original *Overnight* dataset is a valuable data resource for studying semantic parsing as the dataset contains various domains focusing on different linguistic phenomena; the utterances in each domain are annotated both with logical forms (LFs) and canonical forms (CFs). However, as Jia and Liang [2016] point out, this dataset has two main drawbacks: 1) it contains too few entities compared to real datasets, 2) Most of the LFs in test are already seen during training. In consequence, the results achieved on this dataset by different systems [Wang *et al.*, 2015; Jia and Liang, 2016; Xiao *et al.*, 2016] are probably too optimistic.

To remedy these issues, we release an extended *Overnight+* dataset.[4] First, we group all the data and propose a new split. This split makes a 80%-20% random split on all the LFs and keeps the 20% LFs (together with their corresponding utterances) as test and the remaining 80% as training. Thus LFs seen in test are guaranteed to not be seen during training. For each domain, we also add new named entities into the knowledge base and create a new development set and test set containing those new named entities.[5] Depending on the domain, the number of annotated utterances vary from 800 to 4000 and we eliminate some erroneous annotations from the training set. All the reported experiments are conducted on *Overnight+*.

### 4.2 Implementations

For our BRNN, the background $b$ is composed of a WCFG factor ($GW_0$ in subsection 3.2) and depending on the input, zero to several WFSA factors favoring the presence of certain entities. In the current implementation, we only employ automata that have the same topology as the automaton shown in the middle of Fig. 2 where the output vocabulary consists in rule names (e.g. s0, np1) and where the weight $\eta$ is chosen in [0, 0.0001, 0.01] based on the results obtained on the development set.

Currently, we detect only named entities and dates by using mostly exact string matching (e.g. if we detect 'alice' in the input, we construct an automaton to favor its presence in the LF), as well as a small amount of paraphrasing for dates (e.g we detect both 'jan 2' (CF) and 'january 2' as January 2nd). We use a library developed by Wilker Aziz[6] for performing the intersection between WFSA(s) and WCFG. The intersection algorithm results in a new WCFG, from which the background is computed through prefix-conditional probabilities as explained in section 3.

---

[4]https://github.com/chunyangx/overnight_more

[5]We use a high-precision heuristic substituting the named entities in the input string with new ones under certain conditions.

[6]https://github.com/wilkeraziz/pcfg-sampling.

We adopt the same neural network architecture as [Xiao *et al.*, 2016]. We represent the NL semantics by a vector $u_b$ calculated from the concatenation of a vector $u_1$ encoding the sentence at the level of unigrams and another vector $u_2$ at the level of bigrams. Dropout [Srivastava *et al.*, 2014] is applied to $u_1$ (0.1) and $u_2$ (0.3). We model the DS up to time $t$ with the vector $u_t$ generated by an LSTM [Hochreiter and Schmidhuber, 1997]; We concatenate $u_t$ and $u_b$ and pass the concatenated vector to a two-layer MLP for the final prediction. At test time, we use a uniform-cost search algorithm [Russell and Norvig, 2003] to produce the DS with the highest probability. All the models are trained for 30 epochs.

### 4.3 Experimental Results

Table 1 shows the results of different systems. The best average accuracy is achieved by our proposed system BDSP-CL. The system largely improves (48.8% over 34.5% in accuracy) over its RNN baseline DSP-CL which does not have input-dependent WFSA factors. Our system also improves largely over SPO (no-lex) i.e., SPO without "alignment features" (this system still has a rich feature set including string matching, entity recognition, POS tagging, denotation, etc).

In average, BDSP-CL also performs better than the system noted SPO* with the full set of features, but to a more moderate extent. However the results of this SPO* may be too optimistic: the so-called "alignment features" of SPO were obtained from a provided alignment file based on the original *Overnight* training dataset and not on the correct *Overnight+*, because we did not have access to easy means of recomputing this alignment file. The implication is that those features were calculated in a situation where most of the test LFs were already seen in training as explained in 4.1, possibly unfairly helping SPO* on the *Overnight+* test set.

### 4.4 Result Analysis

**Examples** We look into predictions of BDSP-CL, DSP-CL, SPO (no-lex) trying to understand the pros and the cons of our proposed model. Table 2 shows some typical cases. Because our current implementation with automata is limited to taking into account prior knowledge on only named entities and dates, our BDSP-CL can miss some important indications compared to SPO (no-lex). For example, for the sentence 'what locations are the fewest meetings held', our model predicts a set of meetings while SPO (no-lex) detects through its features that the question asks about locations; our model seems better at discovering regularities in the data for which SPO (no-lex) does not have predefined features. For example, for the sentence 'which men are 180cm tall', our model successfully detects that the question asks about males.

Our model consistently performs better than DSP-CL. The example 'what position is shaq oneal' in Table 2 illustrates the difference. In this example, both BDSP-CL and SPO (no-lex) correctly predict the LF; however, DSP-CL fails because it cannot predict the entity 'shaq oneal' as the entity is never seen in training.

**Background Effect** If our background is in average closer to the true distribution compared to the uniform distribution,

| New split | Basketball | Social | Publication | Calendar | Housing | Restaurants | Blocks | Avg |
|---|---|---|---|---|---|---|---|---|
| SPO (no-lex) | 42.2 | 3.1 | **33.1** | **38.8** | 31.0 | **65.4** | 32.1 | 35.1 |
| SPO* | *47.4* | *40.4* | *43.4* | *56.6* | *30.7* | *67.8* | *37.0* | *46.2* |
| DSP-CL | 51.0 | 49.7 | 15.2 | 22.5 | 28.7 | 58.7 | 15.9 | 34.5 |
| BDSP-CL | **63.0** | **57.3** | 25.5 | 36.4 | **60.7** | 64.3 | **34.7** | **48.8** |

Table 1: Test results over all the domains on *Overnight+*. The numbers reported correspond to the proportion of cases in which the predicted LF is interpretable against the KB and returns the correct answer. DSP-CL is the model introduced in [Xiao *et al.*, 2016] that guarantees the grammaticality of the produced DS. BDSP-CL is our model integrating various factors (e.g WCFG, WFSA) into the background. SPO (no-lex) is a feature-based system [Wang *et al.*, 2015] where we desactivate alignment features. SPO* is the full feature-based system but with unrealistic alignment features (explained in 4.3) and thus should be seen as an upper bound of full SPO performance.

| sentence | BDSP-CL | DSP-CL | SPO(no-lex) |
|---|---|---|---|
| 'what locations are the fewest meetings held' | 'meetings that has the least number of locations' | 'location that is location of more than 2 meeting' | *'location that is location of the least number of meeting'* |
| 'which men are 180cm tall' | 'person whose gender is male whose height is 180cm' | 'person whose height is 180cm' | 'person whose height is at least 180cm' |
| 'what position is shaq oneal' | *'position of player shaq oneal'* | 'position of player kobe bryant' | *'position of player shaq oneal'* |

Table 2: Some prediction examples of BDSP-CL, DSP-CL and SPO (no-lex). For readability, instead of showing the predicted LF, we show the equivalent CF. Correct predictions are noted in italics.

|  | DSP-CL | BDSP-CL |
|---|---|---|
| Avg. KL-divergence | 3.13 | 1.95 |

Table 3: Average KL-divergence to the uniform distribution when models predict rules corresponding to named entities.

we hypothesize that the RNN will learn to predict a more uniform distribution compared to an RNN without background as explained in subsection 3.1. To test this hypothesis, we randomly sample 100 distributions in housing domain when the RNN needs to predict a rule corresponding to a named entity. We calculate the average KL-divergence from these distribution to the uniform distribution and report the results in Table 3. The results seem to confirm our hypothesis: the KL-divergence is much smaller for BDSP-CL where a background takes into account the presence of certain named entities depending on the input.

## 5 Related Work and Discussion

Our work makes important extensions over the work [Xiao *et al.*, 2016]. While Xiao et al. [2016] incorporate grammatical constraints into RNN models, we incorporate additional prior knowledge about input dependency. We propose to take into account the well-formedness of LFs by a WCFG and depending on the input, take into account the presence of certain entities inside LFs by WFSA(s). We choose to use WFSA modeling our input-dependent prior knowledge as the algorithm of intersection can efficiently combine WCFG and WFSA(s) to form the background priors guiding the RNN.

Taking into account prior knowledge about named entities is common in more traditional, symbolic semantic parsing systems [Reddy *et al.*, 2014; Berant *et al.*, 2013; Kwiatkowski *et al.*, 2013]. We propose to incorporate those knowledge into an RNN-based model. This is arguably more principled than Jia and Liang [2016]'s approach who incorporate such knowledge into an RNN using data augmentation.

The intersection algorithm used to compute the background allows local weight changes to propagate through the grammar tree thereby influencing the weight of each node inside the tree. This is related to the recent reinforcement learning research for semantic parsing [Liang *et al.*, 2016; Mou *et al.*, 2016] where rewards are propagated over different action steps.

More generally, our work is another instance of incorporating prior knowledge into deep learning models. We do this using symbolic objects such as grammar and automata. In contrast, Salakhutdinov et al. [2013] model prior knowledge over the structure of the problem by combining hierarchical bayesian models and deep models while Hu et al. [2016] handle prior knowledge that can be expressed by first-order logic rules and uses these rules as a teacher network to transfer knowledge into deep models.

## 6 Conclusion and Perspectives

We propose to incorporate a symbolic background prior into RNN based models to learn a semantic parser taking into account prior knowledge about LF well-formedness and about the likelihood of certain entities being present based on the input. We use a variant of a classical dynamic programming intersection algorithm to efficiently combine these factors and show that our Background-RNN yields promising results on *Overnight+*. In the future, we plan to explore the use of WFSA(s) with different topologies to model further prior knowledge.

## Acknowledgements

We thank Wilker Aziz for his help on his intersection algorithm implementation, as well as Matthias Gallé and the anonymous reviewers for their constructive feedbacks.